# Generalized Dynamic Brain Functional Connectivity Based on Random Convolutions

Yongjie Duan, Zhitao Guo, Xi Wu, Zhenjiang Cui, Vince D. Calhoun, Zhiying Long

***Abstract*—Dynamic functional connectivity (DFC) analysis has been widely applied to functional magnetic resonance imaging (fMRI) data to reveal time-varying dynamic changes of brain states. The sliding window method is by far the most popular DFC analysis method due to its simplicity. However, the sliding window method comes with some assumptions, namely the typically approach uses a single window which captures dynamics only within a specific frequency range. In this study, we propose a generalized approach to dynamics via a multi-dimensional random convolution (RandCon) DFC method that is able to effectively capture time-varying DFC at arbitrary time scales by extracting different local features from fMRI time series using a number of multi-dimensional random convolution kernels without the need for learning kernel weights. Compared to a standard sliding window method, multiplication of temporal derivatives (MTD) and phase synchrony methods, RandCon with the smallest kernel size (3 time points) showed notable improvements in performance on simulated data, particularly in terms of DFC temporal and spatial estimation in very short window/kernel size under different noise levels. Results from real fMRI data indicated that RandCon was more sensitive to gender differences than competing methods. Furthermore, we show that the sliding window method can be considered a special case of the proposed multi-dimensional convolution framework. The proposed method is simple and efficient significantly broadens the scope of dynamic functional connectivity research and offer theoretical and practical potential.**

***Index Terms*—Brain state, Convolution, Dynamic functional connectivity, fMRI.**

## I. INTRODUCTION

RESTING-STATE functional magnetic resonance imaging (fMRI) is frequently used to investigate the cooperative functioning of various brain regions and to construct functional connectivity networks[1], [2]. Recent studies indicate that the brain's resting state is dynamic, with functional connectivity (FC) fluctuating over time[3], [4]. According to this, dynamic functional connectivity (DFC) methods have been developed and have advanced significantly in recent years[5]. The dynamic properties of functional connectivity are intricately linked to cognitive processes[6], [7] and cognitive disorders[8], [9].

Various DFC methods including sliding window, time-frequency analysis[10], hidden Markov modeling[11], multiplication of temporal derivatives (MTD)[12], phase synchrony, and other data-driven models[13], [14], [15] have been applied to fMRI data to reveal brain dynamics. Among those DFC methods, the sliding window approach remains the most simple, classical and widely used method[3], [4], [16]. However, a limitation of the approach is the requirement to select one window size[3], [17], where shorter window lengths capture more rapidly changing signals, but can be noisier, while longer window lengths reduce the ability to capture rapid dynamics[17], [18]. As the dynamics are not known a priori, the window size choice may result in a solution which is not well optimized[19].

In recent years, several studies have aimed to refine the sliding window method to improve its capacity for detecting brain dynamics and enhancing interpretability[4], [5]. However, some existing limitations still remain. In each sliding window, the signal amplitude at each time point can be considered as an extracted feature, which results in the feature dimension equaling the window length and the large impact of window length on the time-varying functional connectivity extracted by the sliding window method. Therefore, it is crucial to investigate methods for effectively extracting sufficient features from each ROI's signal without relying on window length.

Convolution has been widely applied to extract local features from signals by sliding a number of convolution kernels over a signal[20]. Multi-dimensional convolution employs multiple convolution kernels, each with distinct parameters, to extract features from the original signal. Although the convolution kernel is similar to a sliding window, the kernel width does not determine the feature dimension. Thus, it is essential to investigate how to incorporate convolution into DFC analysis. In this study, we propose a multi-dimensional random convolution (RandCon) DFC method that is able to effectively capture time-varying DFC even at very short time scales. This method provides a generalized approach to DFC by covering arbitrary window sizes and fully sampling the feature space. First, RandCon is applied to extract different local features from each ROI's time series by setting multi-dimensional convolution kernels with random kernel weights. Based on these extracted features, the time-varying FC matrix is obtained by calculating the correlation of features between pairs of ROIs. Finally, K-means clustering is applied to all the time-varying FC matrices to identify intrinsic brain states. The simulated

This research was funded by the National Natural Science Foundation of China (Grant No. 62071050). *(Corresponding author: Zhiying Long)*

Y. Duan and Z. Cui are with The State Key Laboratory of Cognitive Neuroscience and Learning and IDG/McGovern Institute for Brain Research, Beijing Normal University, 100875 Beijing, China (e-mail: yongjieduan@mail.bnu.edu.cn).

Z. Guo, X. Wu, Z. Long are with the School of Artificial Intelligence, Beijing Normal University, 100875 Beijing, China (e-mail: zlong@bnu.edu.cn).

V. D. Calhoun is with the TReNDS Center, Georgia State University, US 30303 USA.

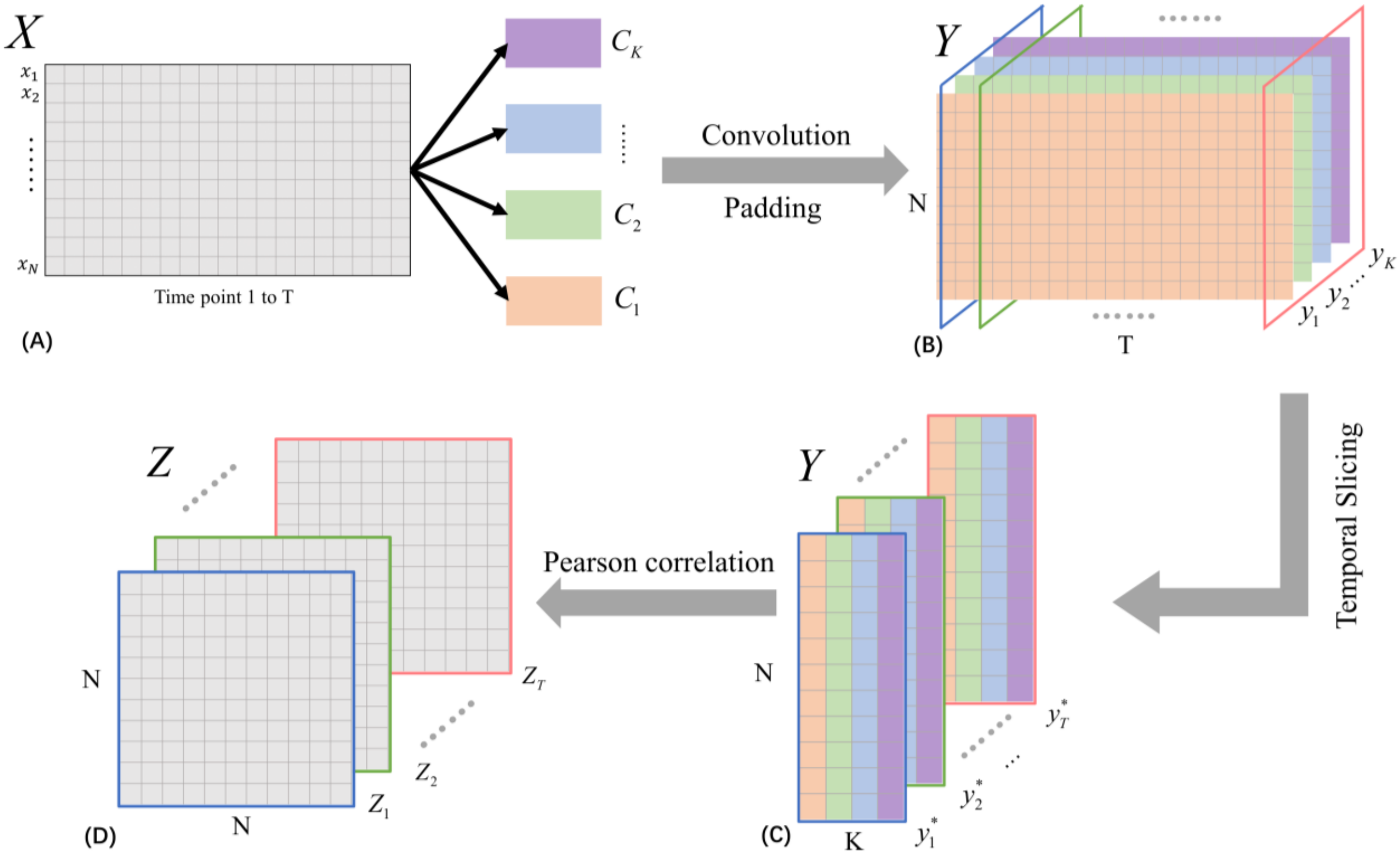

Fig. 1 Overall flowchart of the RandCon method. (A) The time series of N ROIs from a single subject. (B) Feature extraction by convolution. $Y \in \mathbb{R}^{N \times T \times K}$ represents $K$ -dimensional features of $N$ ROI at $T$ time points and $y_k$ corresponds to a feature map derived from filtering the original series using convolution kernel $C_k$ . (C) Temporal slicing to obtain the K-dimension features of N ROIs at each time point. Temporal slicing means to slice the matrix $Y$ into a set of matrices $y_t^*$ along the temporal dimensions, with $y_t^* \in \mathbb{R}^{N \times K}$ representing the K-dimensional features of all ROIs at a specific time point. (D) Time-varying FC matrices calculation. For each $y_t^*$ matrix, the FC matrix $Z_t$ can obtained by calculating Pearson correlation coefficients between $N$ ROIs using $K$ -dimensional features in $y_t^*$ . These $T$ FC matrices are then combined to construct the final result matrix $Z \in \mathbb{R}^{T \times N \times N}$ .

experiments demonstrated the feasibility and robustness of the RandCon DFC method using small kernel size with three time points and proved that RandCon DFC outperformed the sliding window, MTD and phase synchrony in both the temporal and spatial properties of DFC. Meanwhile, the real fMRI experiments demonstrated that the RandCon method was more sensitive to inter-group DFC differences relative to the traditional sliding window method.

More importantly, we show that the multi-dimensional convolution method is a general convolution framework that theoretically includes both RandCon and the sliding window method. In this framework, the sliding window method and its variants are special cases of multi-dimensional convolution by using fixed pulse filter as kernels. The special convolution kernel used in the sliding window method shows significant limitations in window/kernel size and information capacity. By randomly selecting multi-dimensional temporal convolution kernels in RandCon, these limitations can be removed. This study introduces a new approach for dynamic functional connectivity, greatly expanding the exploration space and potential of DFC to study brain dynamics.

## II. Methods

### A. Multi-dimensional Convolution DFC

The flowchart of multi-dimensional convolution is shown in Figure 1. It is assumed that $X = (x_1, x_2, ..., x_N)^T \in \mathbb{R}^{N \times T}$ represents a subject's fMRI signal with $T$ time points from $N$ ROIs, where $x_n \in \mathbb{R}^{1 \times T}$ represents the time series of the $n-th$ ROI (see Figure 1(A)). A total of $K$ convolution kernels $\{C_1, C_2, ..., C_K\}$ are designed to perform convolution on $X$ . When the $k-th$ convolution kernel $C_k \in \mathbb{R}^{1 \times W}$ is applied to convolve with the fMRI time series $x_n(t)$ of the $n-th$ ROI, ultimately producing the $k-th$ feature series $y_k(n,t)$ of the $n-th$ ROI (see equation (1)).

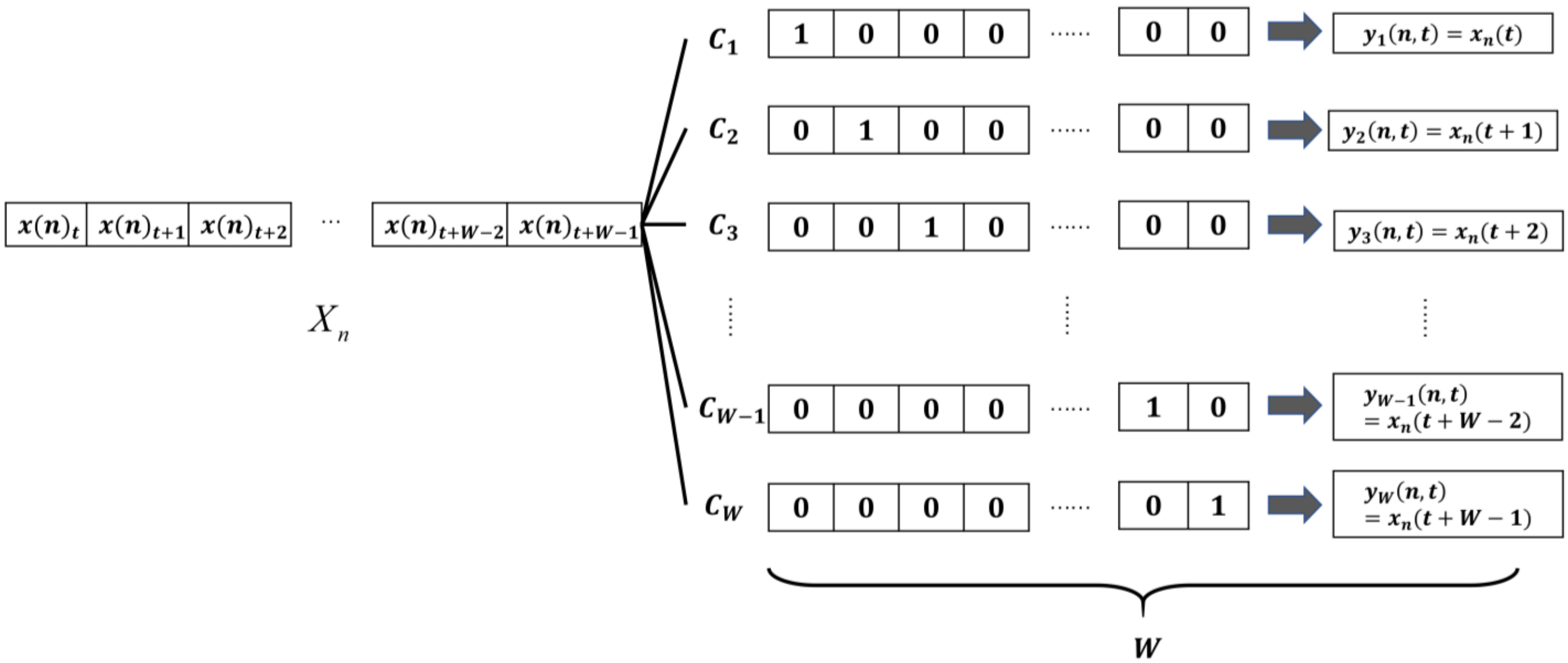

Fig.2 Using a special convolution kernel produces the same results as the sliding window method, demonstrating that the sliding window method is a special case of multi-dimensional convolution. $X$ refers to the data of $W$ time points for a specific ROI of a subject, representing features extracted using a sliding window method with a window length of $W$. After convolution with $W$ specially constructed convolutional kernels, we obtain $d = X$, meaning that this convolution method produces results that are completely consistent with those of the sliding window method.

$$y_k(n,t) = \sum_{w=1}^{W} x_n(t+w-1)C_k(w) \tag{1}$$

Padding is performed to keep the time point number of $y_k(n,t)$ the same as that of $x_n(t)$ by replicating the last time point of $x_n(t)$ $W-1$ times. After convolution with padding, the $k-th$ feature matrix $Y_{\mathrm{k}} \in \mathbb{R}^{N \times T}$ can be obtained. Each row of $Y_k$ contains the $k-th$ feature series of $N$ ROIs. After all the $K$ kernels are applied to all the ROI's fMRI signal, the overall output $Y = \{y_k(n,t)\} \in \mathbb{R}^{K \times N \times T}$ can be obtained (see Figure 1(B)). Next, we slice the output $Y$ based on the temporal dimension by setting $Y = \{y_t^*(n,k)\} \in \mathbb{R}^{K \times N \times T}$ to obtain the feature matrix at each time point, where $y_t^*(n,k) = Y(k,n,t)$. Each row of the feature matrix at the $t-th$ time point $y_t^* \in \mathbb{R}^{N \times K}$ represents each ROI's $K$-dimensional features that are extracted through multi-dimensional convolution at time point $t$ (see Figure 1(C)). For the $m-th$ and $n-th$ ROIs, the FC between the two ROIs at time point t can be calculated by using the Pearson correlation in equation (2), thereby forming a time-varying FC matrix $Z_{\mathrm{t}}$.

$$Z_t(n,m) = Z_t(m,n) = \frac{\sum_{k=1}^{K}(y_t^*(m,k) - \overline{y_t^*(m)})(y_t^*(n,k) - \overline{y_t^*(n)})}{\sqrt{\sum_{k=1}^{K}(y_t^*(m,k) - \overline{y_t^*(m)})^2}\sqrt{\sum_{k=1}^{K}(y_t^*(n,k) - \overline{y_t^*(n)})^2}} \tag{2}$$

Where $\overline{y_t^*(m)} = \frac{1}{K}\sum_{k=1}^{K} y_t^*(m,k)$, $\overline{y_t^*(n)} = \frac{1}{K}\sum_{k=1}^{K} y_t^*(n,k)$.

Here, $Z_t \in \mathbb{R}^{N \times N}$ represents the FC matrix at time point $t$. All diagonal values of the symmetric matrix $Z_t$ are equal to 1. To remove redundant information, the diagonal values are removed and only the lower half of the matrix $Z_t$ are maintained, resulting in a time-varying FC vector $V_t$ with length $\frac{N \times (N-1)}{2}$. The K-means clustering method is used to group the time varying FC vectors of all the $T$ time into $M$ cluster centers with $\frac{N \times (N-1)}{2}$ dimensions. At clustering, each time point is assigned to a cluster. A state time series can be obtained by arranging the cluster label of each time points as a time sequence.

### B. Sliding Window Method as a special case of Multi-dimensional Convolution

From the perspective of multi-dimensional convolution, the sliding window method can be regarded as a special case of multi-dimensional convolution by setting each convolution kernel as a pulse response function. As depicted in Figure 2, it is assumed that there are $W$ convolution kernels $\{C_1, C_2, ..., C_W\}, C_w \in \mathbb{R}^{1 \times W}$, where the $w-th$ kernel has a parameter of 1 at the $w-th$ position and 0 elsewhere. On the framework of multi-dimensional convolution, the first feature series $y_1(n,t)$ can be extracted from the time series $x_n$ of the $n-th$ ROI by employing the $1-th$ convolution kernel $C_1$ (see equation (3)).

$$y_1(n,t)=\sum_{w=1}^{W}x_n(t+w-1)C_1(w)=x_n(t) \quad (3)$$

According to equation (3),

$$y_1(n,t)=x_n(t),\ y_2(n,t)=x_n(t+1),...,\ y_W(n,t)=x_n(t+W-1) \quad (4)$$

Suppose the vector $d_n(t)$ represents the features that are extracted from the $n-th$ ROI by the $W$ convolution kernels with the form in equation (4) at the $t-th$ time point. The feature vector $d_n(t)$ can be expressed by equation (5).

$$d_n(t)=[y_1(n,t),y_2(n,t),...,y_W(n,t)]= [x_n(t),x_n(t+1),...,x_n(t+W-1)] \quad (5)$$

It can be seen that the feature vector $d_n(t)$ contains $W$ time points that are truncated from the time series $x_n$ by a sliding window with the window length equal to $W$. Therefore, it is proved that the sliding window method is a special type of multi-dimensional convolution method.

## III. Experiment

### A. Analysis of Simulated Dataset

#### 1) Simulated Data Generation

The simulated experiments were performed to evaluate the effectiveness of the RandCon method. The osl-dynamics toolkit [21](https://github.com/OHBA-analysis/osl-dynamics) was employed to generate simulated fMRI time series. First, we assumed that there existed four intrinsic brain states in the simulated data and used a random algorithm to generate states as "ground truth". To establish certain group relationships between ROIs in each brain state, we divided all ROIs into some groups. Each group consisted of 10 ROIs, as shown in Figure 3, where the smallest grid represented one group. For the 10 ROIs within each group, we maintained a completely consistent synchronization relationship at all times. Each pair of ROIs within each group was assumed be to highly positive correlation with correlation coefficient equal to 1. The relationship between groups was randomly set to be completely negative correlation (correlation coefficient = -1), unrelated (correlation coefficient = 0), and completely positive correlation (correlation coefficient =1). The duration of each state was determined by a Gamma distribution with a shape parameter and a scale parameter. The shape parameter was set to 10 in this study, and the scale parameter was set to 5. For each combination of parameters (time point number and ROI number), simulated data of 600 subjects were generated as a baseline dataset without noises. Within each dataset, the ground truth of latent states remained consistent across these 600 subjects. Figure 3 displays the four states generated by osl-dynamics toolkit, along with the state time series of the first subject.

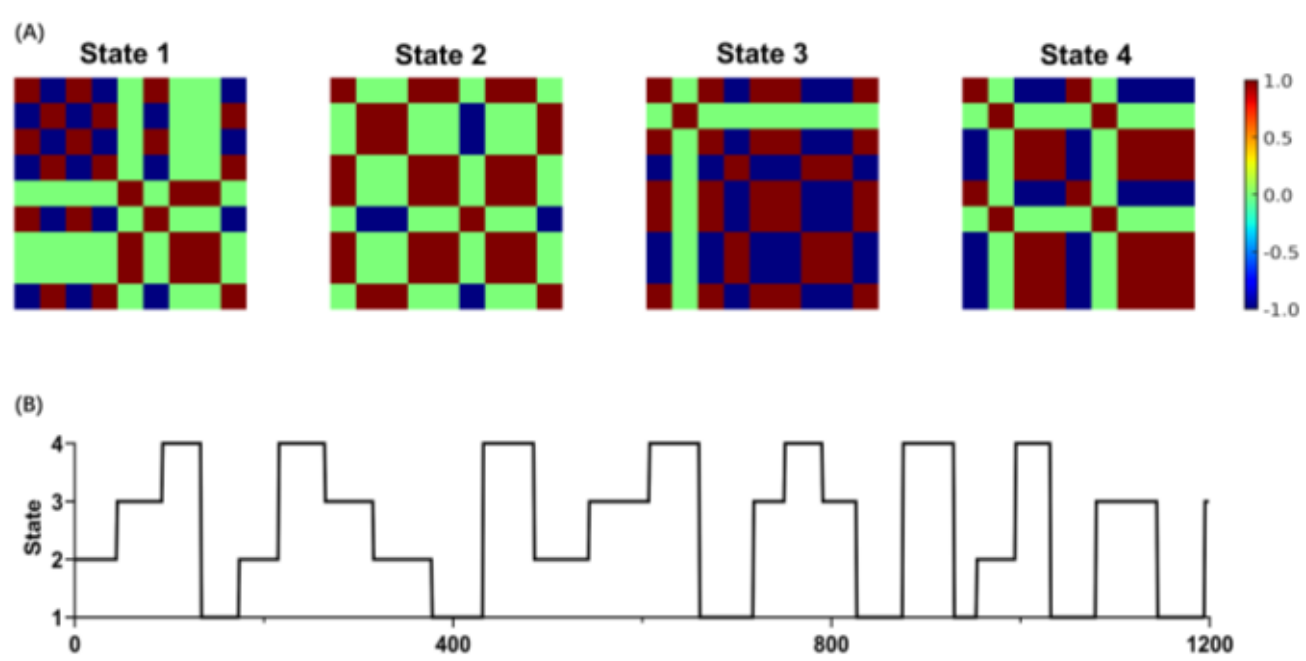

Fig.3 Simulated data generated using osl-dynamics, along with the state sequences of the first subject. (A) Generated ground truths of the four states' spatial patterns. (B) State time series of the generated states of the first subject.

#### 2) Validation of robustness

In the simulated data experiment, we investigated the robustness of the RandCon DFC method and compared the performance of RandCon with the MTD, phase synchrony and sliding window DFC methods. The MTD method was implemented as described in the original paper[12], using the provided code in the previous study. After obtaining the functional connectivity matrix at each time point through MTD, a moving average was applied, with the window length =3 and step size=1 consistent with those used in the sliding window and RandCon methods. The phase synchrony method followed the approach by Yaesoubi[22], which involved performing a Hilbert transform on the BOLD signal and calculating the functional connectivity matrix using the instantaneous phase. The sliding window method employed the classic approach[10], where a segment of the signal within the window was extracted with the window size =3, and the Pearson correlations between ROIs were calculated to obtain the time-varying FC matrix. The RandCon method derived the time-varying FC matrix using equation (2), further details are provided in the Methods section. Both the sliding window length and the convolution kernel length were set to 3. For RandCon, each parameter of each convolution kernel was sampled uniformly from a standard Gaussian distribution without further training or adjustment in this study. Except for Section 3.1, the kernel number is set to 2048 in all other simulated data experiments.

#### 2.1) Robustness to Noise

Gaussian Noise with a zero mean and a specific standard deviation (SD) was added to a baseline dataset including 600 subjects to generate a simulated dataset with a noise level. The parameters for the baseline dataset were set as time point number=1200, ROI number=90. Seven simulated noise datasets were generated by varying noise SD from 0.4 to 1 increasing by 0.1 increments. For each dataset, the 600 subjects were randomly divided into 30 groups, each containing 20 subjects. The RandCon, sliding window, MTD and phase synchrony were applied to each subject's data to obtain the time-varying FC matrices separately. For each method, all the time-varying FC matrices of the 20 subjects within each group were aggregated and inputted to the K-means clustering. K-means clustering was performed 100 times on each input, with each clustering undergoing 20 iterations. From these 100 times of clustering, we identified the best clustering result using the Davies-Bouldin Index[23] as the evaluation criterion. For each subject, the Adjusted Rand Index (ARI) that measured the accuracy of the estimated state time series and the cosine similarity that measured the accuracy of the estimated FC patterns were calculated. For each group at a specific noise level, the mean ARI and cosine similarity were calculated by averaging the results across 20 subjects. The two-way ANOVA tests were applied to the mean ARI and cosine similarity of 30 groups at each noise level to detect the differences among the four methods. The significance levels of all the tests were set to $p<0.05$.

*2.2) Robustness to ROI Numbers*

To compare the impact of the number of ROIs on the method's performance, we fixed the noise level at 0.6, and varied the number of ROIs to 30, 60, 90, and 120 in this experiment. The other parameters remained consistent with those specified in Section 2.1. We generated four datasets corresponding to four ROI numbers separately. Each dataset contained 30 groups and each group consisted of 20 subjects. The four DFC methods were applied to all the datasets to extract the intrinsic brain states and state time series. The ARI and the cosine similarity values of each subject were calculated. For each group with a specific ROI number, the mean ARI and cosine similarity were calculated by averaging the results across 20 subjects. Two-way ANOVA analysis with the Bonferroni correction method were applied to the mean ARI and cosine similarity of 30 groups to detect whether there were significant differences among the four methods for each ROI number. The significance levels of all the tests were set to $p<0.05$.

*2.3) Robustness to Number of Time Points*

In order to investigate the impact of time length of time series on the four DFC methods, we generated five simulated datasets with the noise SD equal to 0.6. The time point number of the five datasets was set to 400, 600, 800, 1000, and 1200 separately. The other parameters remain unchanged with Section 2.1. All subsequent steps after the dataset generation are consistent with those described in Section 2.2.

*3) The Impact of Kernel Size and Kernel Number on RandCon*

Although the kernel weights of RandCon can be set randomly, it still includes two important parameters that are the size of the convolution kernel and the number of convolution kernels. The size of the convolution kernel determines the number of time points considered in feature extraction during convolution, while the number of convolution kernels determines the dimensionality of the extracted features. In the following sections, we investigated how these two parameters influenced RandCon's performance. We utilized the dataset generated as described in Section 2.1.

*3.1) Number of Convolution Kernels*

The number of convolution kernels is a key parameter that affects the feature dimensionality and noise robustness of RandCon. In this experiment, we evaluated the impact of the number of convolution kernels on the method's performance. Four simulated datasets with noise SD equal to 0.6, 1.1, 1.6 and 2.1 were generated in the same way as the experiment in section 2.1. RandCon with kernel number equal to 5, 10, 20, 40, 80, 160 was applied to the simulated data of each subject separately. The mean ARI and cosine similarity of each group were obtained by averaging individual values across 20 subjects. Finally, the mean ARI and cosine similarity of each noise level were calculated by averaging the mean values of individual group across the 30 groups.

*3.2) Size of Convolution Kernels*

In this experiment, we evaluated the impact of convolution kernel size/window size on RandCon/sliding window. Large convolution kernel needs more paddings to ensure the temporal dimension unchanged after temporal convolution. Because excessive padding can lead to distortion, we did not use padding for any kernel size in this experiment. The simulated dataset with a noise SD equal to 0.6 generated in Section 2.1 was used. Both RandCon and sliding window methods with kernel/window size equal to 4, 5, 6, 7, 8, and 9 were applied to the simulated dataset separately. Because convolution without padding resulted in the reduction of time points, the KL divergence that measures the discrepancy between the distribution of the estimated state dwell time and the distribution of the ground truth was used to measure the accuracy of state time series. The state dwell time represents the mean duration of each state. Both cosine similarity and mean square error (MSE) were used to measure the accuracy of the estimated state pattern. For each kernel/window size, the mean KL divergence, cosine similarity and MSE of each group were calculated by averaging the individual values across 20 subjects. The two-way ANOVA tests were applied to these metrics of 30 groups to detect the differences between the two methods for each window/kernel size. The significance levels of all the tests were set to $p<0.05$.

## B. Real fMRI Data Analysis

*1) Data Acquisition and Preprocessing*

Data Acquisition: The Human Brain Connectome Project (HCP) has publicly released resting-state fMRI data from 1200 healthy subjects (https://www.humanconnectome.org/). The rsfMRI data of 100 subjects (age: 22–36 years old, 46 men and 54 women) in this dataset were used in this experiment. Each subject underwent two approximately 15-minute resting-state scans, with eyes open and relaxed fixation on a projected bright cross-hair on a dark background. Whole-brain gradient-echo-planar imaging acquisitions were acquired with the following main parameters: repetition time (TR) = 720 ms, echo time (TE) = 33.1 ms, flip angle = 52°, field of view (FOV) = 208×180 $mm^2$, matrix = 104×90, slice thickness = 2 mm, number of slices = 72, voxel size = 2×2×2 $mm^3$, multiband factor = 8. In this study, the rs-fMRI data of the “LR” encoded run in the first scan (i.e., REST1) were used. For information about data acquisition, please refer to the source study. The ICA-FIX rs-fMRI data of the HCP that produced the “minimal preprocessing pipelines” and ICA-FIX were used in the study[24].

*2) ROI Time Course Extraction*

Ninety ROIs were selected based on the anatomical automatic labeling (AAL) atlas[25], and their time courses were extracted from the preprocessed rsfMRI data using the python toolbox NiBabel (https://nipy.org/nibabel/). Each ROI's time series was normalized to have a mean of 0 and a standard deviation of 1. According to previous studies[26], these 90 ROIs were divided into nine subnetworks: sensorimotor network (SMN), cingulo-opercular network (CON), auditory network (AUN), DMN, visual network (VN), frontoparietal network (FPN), SN, subcortical network (SCN), and none.

*3) Intrinsic Brain State Extraction*

For each subject, both the sliding window method and the RandCon method were applied to all the ROIs' time series to extract time-varying FC patterns. The sliding window method utilized a window of 3 time points with a step size of 1 time point. the RandCon method set the convolution kernel width to 3 and the kernel number to 2048. No padding was performed for the two methods. A total of 1198 time-varying FC matrices with size 90×90 was obtained for each subject. Since each connectivity matrix is symmetric, only the lower triangular parts of 1198 matrices of all subjects were used in the following K-means clustering step. The optimal number of clusters was estimated by using the K-means elbow method[27].

Unlike simulated data, real data encompasses greater and more complex inter-subject variability, noise, and various other factors[5]. To enhance the stability of the clustering results, we employed a stacked denoising autoencoder (SDAE)[28] model for dimensionality reduction on the input data of k-means. The SDAE model maintained the same learning rate, structure, and embedding dimensions as the original paper. We inputted the time-varying FC matrices derived from the sliding window method and the RandCon method into the SDAE model separately, training for 200 epochs. The trained model parameters were then used to perform dimensionality reduction on the connectivity matrices, reducing each FC matrix to 10 dimensions. Subsequently, we applied the K-means clustering method to cluster the reduced FC matrices of all time points from all subjects collectively. The Davies-Bouldin Index was used to evaluate the clustering results, identifying the best clustering outcome out of 100 attempts. For each state, the FC pattern was obtained by averaging the FC matrices whose time points was clustered to the same state.

*4) Temporal and spatial DFC analysis for sliding window and RandCon*

To investigate the temporal dynamic characteristics of the brain states extracted by the sliding window and the RandCon DFC methods, the mean dwell time and the fraction of time were calculated for each state of each subject. The mean dwell time of state k represents the average number of consecutive time points in which state k was active. The fraction of time of state k indicates the proportion of time points in which state k was active.

FC variability and centroid similarity were calculated to measure the spatial dynamic characteristics of each subject. For each subject, the FC variability matrix was obtained by calculating the standard deviation of each connection across all sliding windows. The FC variability of each subject was then calculated by averaging the variabilities across all connections in the upper triangle of the FC variability matrix. For FC centroid calculation, the Fisher transformation was first applied to the FC matrices of each subject, and a mean FC matrix was obtained by averaging all the time-varying matrices across time points. Then, the cosine similarity between the FC matrix at each time point and the mean FC matrix was calculated. The mean FC cosine similarity of each subject was obtained by averaging the centroid similarity across all the time points. The independent two-sample t tests were applied to the FC variability and cosine similarity of the two groups to detect the inter-group differences. The significance level was set to $p<0.05$.

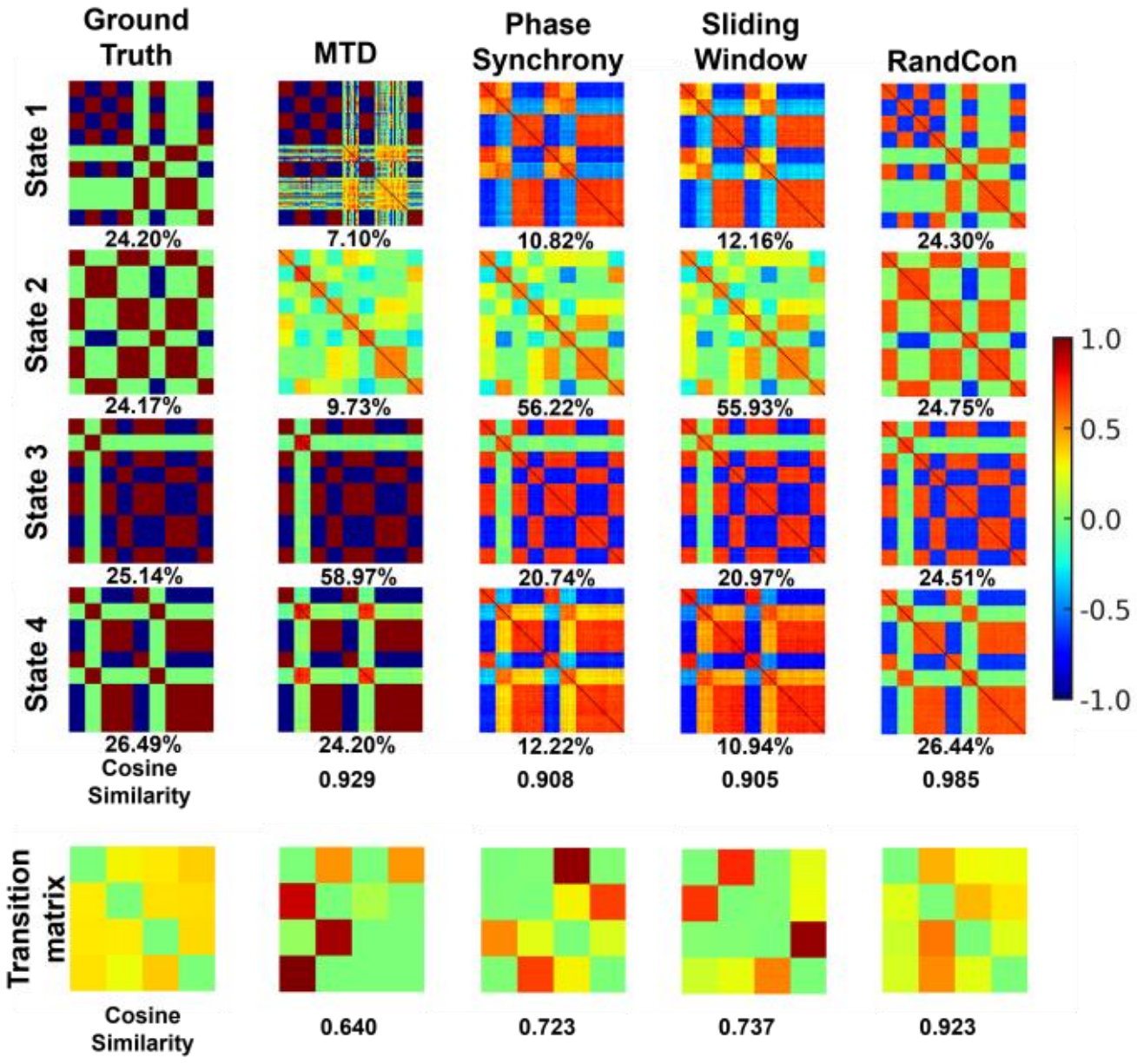

Fig.4 The ground truth of the four states (row 1-4 and column 1) and the four state patterns (row 1-4) and the transition matrices (row 5) that were estimated by MTD (column 2), phase synchrony (column 3), sliding window (column 4) and RandCon (column 5). The percent value below each state represents the proportion of time spent in each state. For the state patterns, the mean cosine similarity of the four states measures the similarity between the estimated states and ground truth. For the state transition matrices, the $n-th$ row represents the transition frequencies from state $n$ to the other states and the cosine similarity measures the similarity between the estimated transition matrices and the ground truth transition matrix. The results are from the first group of 30 subjects, with a noise standard deviation of 0.6, 90 ROIs, and 1200 time points. The kernel/window size is both 3 time points.

## IV. Results

### A. Analysis of Simulated Dataset

*Robustness Against Noise*

Figure 4 shows the four brain state patterns that were estimated by the four DFC methods for a group of 600 simulated subjects at one noise level (standard deviation=0.6). The fraction of time of the four states that were estimated by RandCon was closer to the ground truth than the other three methods. For the spatial pattern of the four states, RandCon showed larger cosine similarity between the estimated states and the ground truth than the other three methods. Furthermore, we counted the number of transitions between states and converted it into frequency, thus obtaining the state transition matrix for each method. As depicted in Figure 4, the cosine similarity between the transition matrix of RandCon and the ground truth is the highest, which indicates that the RandCon method captures the transitions between states very well.

Figure 5(A) presents the mean ARI values of the MTD, phase synchrony, sliding window and RandCon DFC methods under various noise levels. As illustrated, the performance of all methods deteriorated and MTD had the minimum variation as noise variance increases. Among the four methods, RandCon showed the highest ARI values while the MTD method showed the lowest ARI values for all noise levels. The sliding window method and the phase synchrony method showed no significant differences in ARI value. In contrast to the other three methods, RandCon showed significantly higher ARI values at all noise levels ($p<0.001$), indicating that RandCon significantly outperformed the other three methods in state time series estimation.

Figure 5(D) displays the cosine similarity results of the brain states for the four methods under different noise conditions. As noise variance increased, the overall cosine similarity of all methods decreased. The cosine similarity of the RandCon method is significantly higher than those of the other three methods at all the noise levels ($p<0.05$). Moreover, MTD showed significantly larger cosine similarity compared to the sliding window and phase synchrony methods. The cosine similarities of the sliding window and phase synchrony method did not show significant differences. The detailed tables of statistical test results can be found in the Supplementary materials.

*Robustness to ROI Numbers*

Figure 5(B) presents the ARI values of the four methods for different ROI numbers. RandCon exhibited significantly higher ARI values for all ROI numbers than the other methods. Moreover, the ARI values of RandCon increased rapidly with the increasing of ROI number. The sliding window method and the phase synchrony method showed increased ARI value when the ROI number increased from 30 to 60 and remained almost stable for ROI number larger than 60. MTD produced the smallest ARI values among the four methods and showed slight increases in ARI value with increasing ROI number.

Figure 5(E) illustrates the cosine similarity of the four brain states estimated by the four methods for different ROI numbers. For ROI number equal to 30, RandCon and MTD showed significantly higher cosine similarity than sliding window and phase synchrony methods. When the ROI number was larger than 30, the cosine similarity of RandCon was significantly higher than that of the other three methods. Moreover, RandCon exhibited an increased trend with the increasing of ROI number while the other three methods fluctuated with the increasing of ROI number.

*Robustness to Number of Time Points*

Figure 5(C) presents the ARI values of the four methods for different numbers of time points. With an increasing number of time points, the ARI values of the RandCon method improves progressively, and significantly outperformed the other three methods. The MTD method consistently showed the lowest ARI values in all cases while the sliding window method and phase synchrony method remained similar.

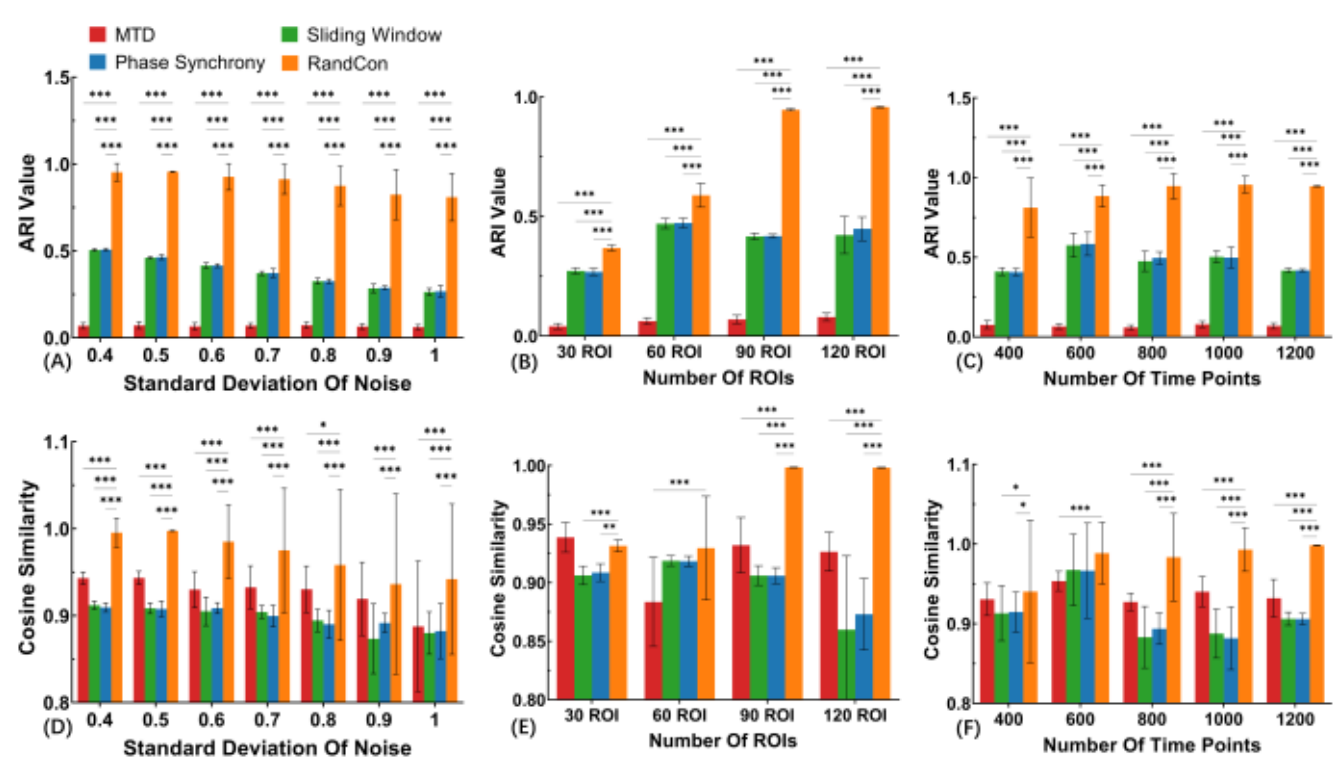

Fig.5 Validation results of simulated data under different noise levels, numbers of ROIs, and numbers of time points for the four methods. (A-C) ARI comparison of the four methods for different noise levels (A), ROI numbers (B) and time point numbers (C). (D-F) Cosine similarity comparison of the four methods for different noise levels (D), ROI numbers (E) and time point numbers (F). The figure only presents the significance results of the RandCon method in comparison to the other three methods and does not display the significance test results of the other three methods (* for $P<0.05$, ** for $P<0.01$, *** for $P<0.001$).

Figure 5(F) illustrates the cosine similarity results of four methods for different numbers of time points. When the number of time points was low, all four methods yield similar results. Specifically, when the number of time points was 400 or 600, the cosine similarity value of RandCon was only significantly higher than some of other methods (sliding window method and phase synchrony method in 400 time points, and MTD in 600 time points). However, when the number of time points was larger than 800, the RandCon method's values improved significantly, resulting in higher performance compared to the other three methods. The results of the other three methods fluctuated with changes in the number of time points.

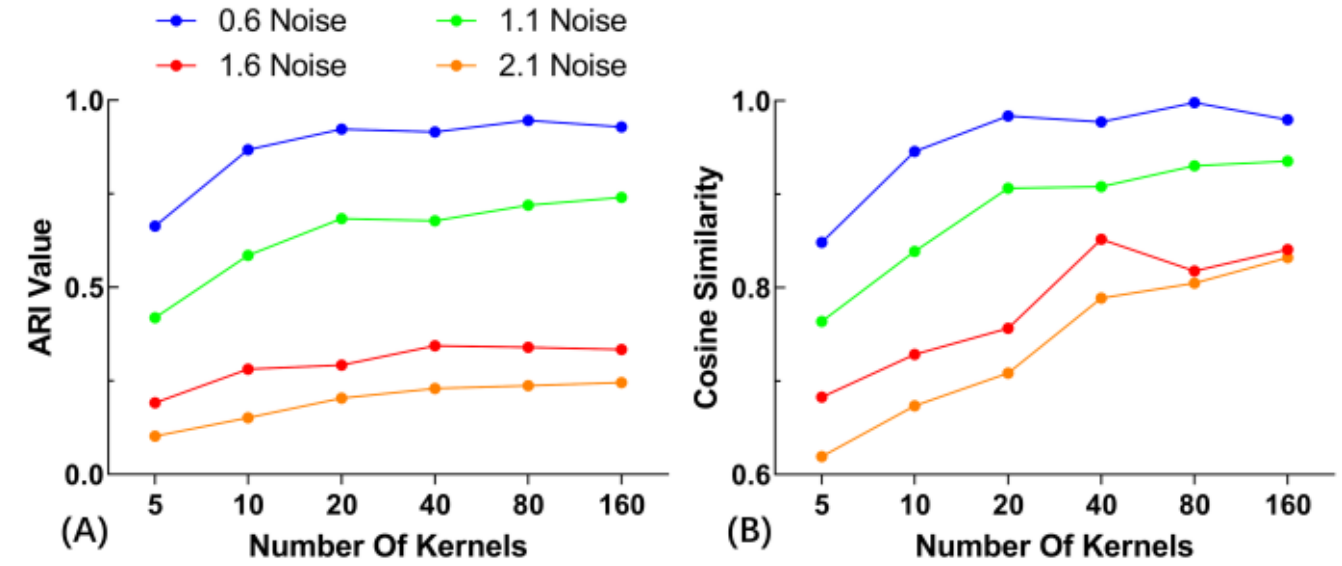

Fig.6 Results of impact of kernel number on RandCon. (A-B) Variation of ARI value (A) and cosine similarity (B) of the simulated data with different numbers of convolution kernels.

*Evaluation of Number of Convolution Kernels*

Figure 6 illustrates the effect of varying the number of convolutional kernels on ARI and cosine similarity of RandCon under different noise conditions. As shown, the overall performance of the RandCon method decreased as noise levels increased for both ARI and cosine similarity. However, within each noise condition, method performance improved with an increasing number of convolutional kernels, indicating that more kernels enhanced feature extraction and noise resistance.

Moreover, when the noise level was low (standard deviation=0.6), the cosine similarity increased with the increasing of kernel number before 20 kernels. However, the increasing trend was not obvious after 20 kernels. In contrast, method performance continued to improve with the increasing of kernel number after 20 kernels at high noise levels (see Figure 6B), which possibly suggested that higher noise levels require more convolutional kernels to achieve asymptotic state.

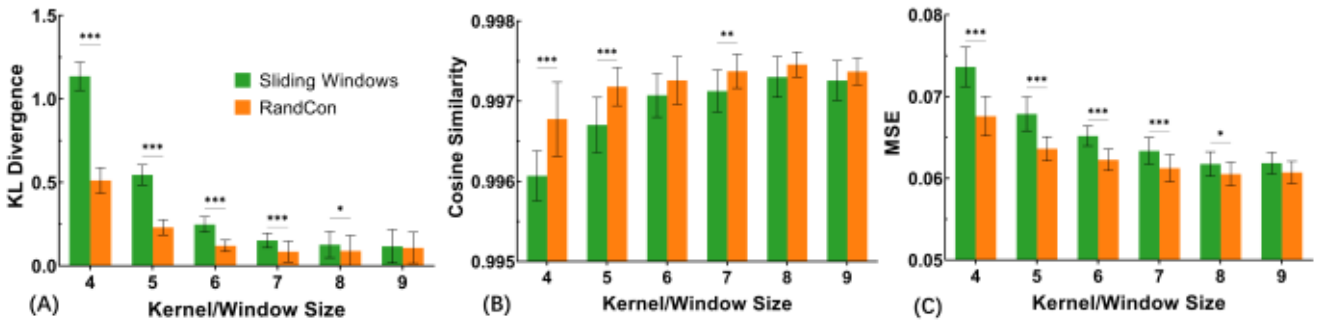

Fig.7 Results of the impact of kernel/window size on RandCon/sliding window. (A-C) The KL divergence (A), cosine similarity (B) and MSE (C) of the two methods for different kernel/window size. (* for P<0.05, ** for P<0.01, *** for P<0.001)

*Evaluation of Convolution Kernel Size*

Figure 7 compares the sliding window method and the RandCon method using different convolutional kernel sizes. Measured by the dwell time, the data showed that as the convolutional kernel/window length increased, the KL divergence decreased for RandCon/sliding window, indicating the distribution of the dwell time estimated by the two methods was closer to that of the ground truth in the case of larger kernel/window size. For the state FC patterns estimated by RandCon/sliding window, the cosine similarity increased and the MSE decreased with the increasing of kernel/window size. Moreover, the RandCon method outperforms the sliding window method in all cases and the performance differences between the two methods decreased with the increasing of kernel/window size.

## B. Real fMRI experiment

The optimal number of brain states was set to be 4 by using the Elbow method[27] for both sliding window and RandCon methods, as shown in Supplementary materials. Figure 8 illustrates the functional connectivity patterns of four intrinsic brain states identified by RandCon and the sliding window methods. The four brain state patterns of RandCon were similar to those of the sliding window method. Among these states, state 3 exhibited overall weak connections, while state 4 exhibited overall strong connections.

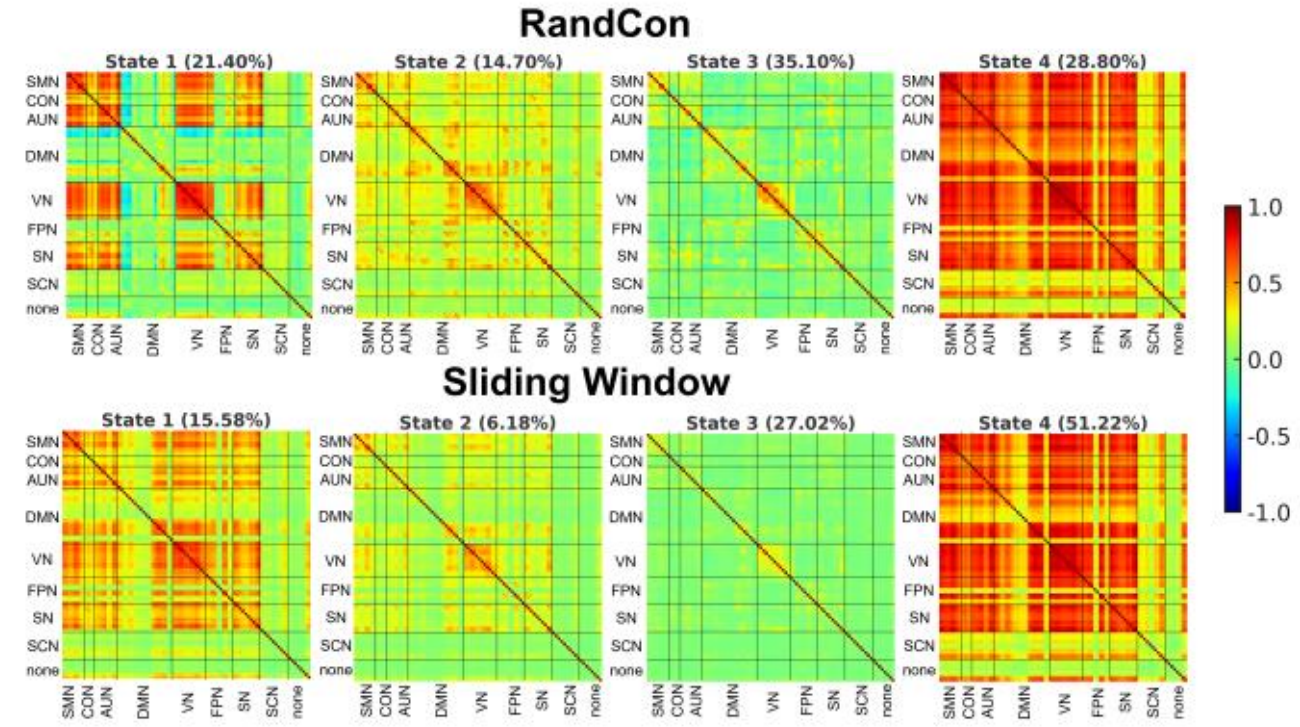

Fig.8 The FC patterns of the four states extracted from real data of one hundred subjects using the RandCon method and the sliding window method, respectively. Nine subnetworks: sensorimotor network (SMN), cingulo-opercular network (CON), auditory network (AUN), DMN, visual network (VN), frontoparietal network (FPN), SN, subcortical network (SCN), and none.

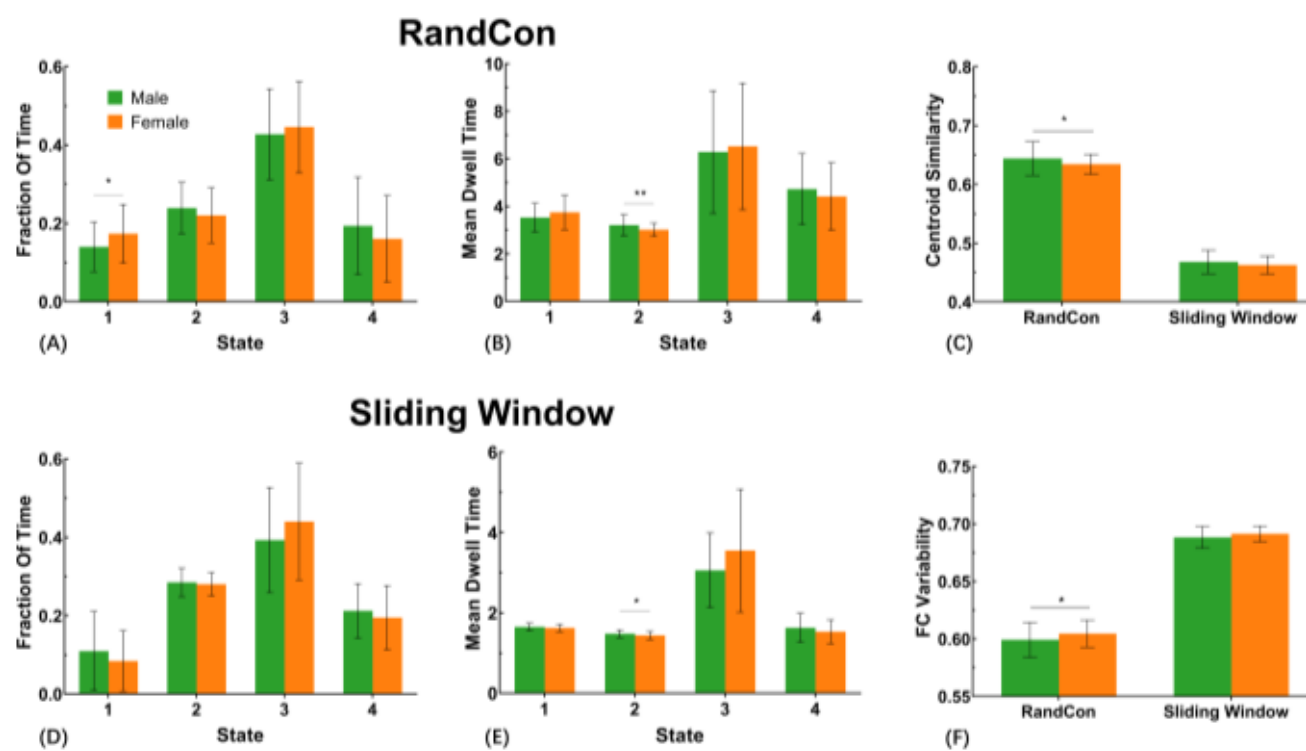

Fig.9 DFC results of the RandCon and sliding window methods. (A-C) The mean fraction of time (A), mean dwell time (B) and mean centroid similarity (C) of the four states estimated by RandCon. (D-F) The mean fraction of time (D), mean dwell time (E) and mean centroid similarity (F) of the four states estimated by the sliding window method. (* for P<0.05, ** for P<0.01, *** for P<0.001)

*Temporal DFC analysis*

The temporal DFC results including the fraction of time and the mean dwell time of each brain state estimated by the two methods are presented in Figure 9. For the fraction of time, RandCon detected that the proportion of time spent by male subjects was significantly lower than that of female subjects in state 1 (see Figure 9(A)) while the sliding window method did not detect any significant gender differences across all states (see Figure 9(D)).

Regarding the mean dwell time, both methods found that the male group tended to stay more time in state 2 compared to the female group (see Figure 9(B) and (E)). In contrast to the sliding window method, RandCon detected larger and more significant inter-group differences ($p<0.01$ for RandCon, $p<0.05$ for sliding window) in state 2.

*Spatial DFC analysis*

The spatial DFC results including FC variability and centroid similarity of the two methods are presented in Figure 9(C) and 9(F). For the RandCon method, the male subjects exhibited significantly higher centroid similarity and significantly lower variability than the female subjects. In contrast, the sliding window method did not detect significant inter-group differences in centroid similarity and FC variability.

## V. Discussion

In this study, we proposed a multi-dimensional RandCon DFC method that employed random multi-dimensional convolution kernels to extract signal features and computed time-varying DFC matrices. Results of simulated data demonstrated that RandCon showed better robustness to noises, the ROI number and the time series length in DFC temporal and spatial estimation than the other three methods. The real resting fMRI data demonstrated that RandCon outperformed the sliding window method in detecting the inter-group differences of DFC temporal and spatial properties. Additionally, we introduced a framework of multi-dimensional convolution that incorporates the RandCon, sliding window method and its variants into the broader theoretical context, enhancing the generality and flexibility of the sliding window DFC method.

For DFC method, it is important to extract different time-varying features from each ROI so that time-varying FC between pairs of ROIs can be obtained. RandCon is a type of multi-dimensional convolution method by sampling each kernel's weights randomly from Gaussian distribution without training, greatly enhancing computational speed. Although this method incorporates randomness, the kernels were applied in the same order and the extracted features were in the same order for all the ROIs. Thus, the correlation between two ROIs is meaningful by using the features extracted by RandCon. Additionally, RandCon allows for flexible kernel number, facilitating flexible changes of feature dimensions.

The validation results from the simulated dataset indicated that the RandCon method consistently surpassed the performance of the other three methods in the ARI values and cosine similarity and showed good stability irrespective of variations in the data generation parameters (see Figures 4 and 5). This clearly demonstrated the stability and robust capabilities of the RandCon method. Experiments on the impact of convolution kernel number, as seen in Figures 6(A) and 6(B), showed that in contrast to the sliding window method, the RandCon method can incorporate multiple feature dimensions in one analysis by utilizing varying numbers of convolution kernels. For the sliding window method, a shorter window size will lead to lower feature dimensions, which will make it difficult to discern whether signal fluctuations arise from meaningful brain activity or from noise. Conversely, the RandCon method can address this issue by raising the feature dimensions. Different convolution kernels filter the original signal in various ways and may extract features from different aspects, significantly enhancing the noise resistance and stability of the features. Additionally, as shown in Figure 7, when the kernel/window size increased, RandCon's performance still significantly surpassed that of the sliding window method in most cases, indicating that the RandCon method was suitable for both shorter and longer kernel/window size. For short convolutional kernels, RandCon's performance advantage was pronounced.

The analysis results of the real dataset further confirmed the effectiveness of the RandCon method. By analyzing rs-fMRI data from 100 subjects, RandCon identified brain states similar to those detected by the sliding window method (see Figure 8). In contrast to the sliding window method, RandCon detected more distinctive DFC state patterns that can provide additional insights into understanding each brain state. Thus, these results indirectly supported RandCon's superiority in brain state space detection capability compared to the sliding window method in simulated experiments.

For state 1, strong positive correlations were observed between SMN, CON, AUN, VN, and SN. SMN could serve as an integrator in the brain and process sensory inputs to form common experiences. Thus, state 1 may be associated with the integration of auditory and visual stimuli and the interpretation of internal and external experiences[29], [30]. In state 2, the positive connectives between the superior medial frontal gyrus and precuneus in DMN and positive connections in VN possibly suggested that subject was engaged in mental imagery of the events that were relevant to self and retrieve from episodic memory with strong visual experience[31], [32]. State 3 shows overall weaker connections than other states and may be associated with relaxation[10], [33], [34]. For state 4, most connections display strong positive correlations. The strong positive connections between DMN, FPN, SN, AUN, VN in state 4 could suggest that subjects switched between internally directed cognition and externally detection and integration of sensory input stimuli in this state.

For the DFC analysis of real fMRI data, both RandCon and the sliding window method detected similar inter-group differences in temporal and spatial DFC measurements. However, RandCon identified more inter-group differences compared to the sliding window method. These results highlighted RandCon's advantage in detecting inter-group DFC differences. We primarily interpreted RandCon's results in the analysis of real fMRI data as follows. Compared to the male group, the female group shows a significantly higher proportion of time in state 1 and a significantly lower average dwell time in state 2 (see Figure 9). Higher fraction of time in state 1 and lower mean dwell time in state 2 of female group may indicate that the female subjects tended to spend more time in detecting external sensory stimuli and less time in internally directed cognition than the male group[35]. From a spatial pattern perspective, the RandCon method found that the central similarity in the male group was significantly higher than that in the female group, which could indicate that the states of the male group at different times are closer to the average state. Additionally, the male group exhibited significantly lower variability than the female group for FC variability, indicating that men's states fluctuated and changed less over time. Both higher central similarity and lower FC variability of males suggested greater state stability in men compared to women. Because female subjects tend to experience more emotional fluctuations compared to male subjects[36], which could result in more state fluctuations in the female group compared to the male group.